\documentclass[letterpaper]{article} 
\usepackage{aaai23}  
\usepackage{times}  
\usepackage{helvet}  
\usepackage{courier}  
\usepackage[hyphens]{url}  
\usepackage{graphicx} 
\usepackage{natbib}  
\usepackage{caption} 
\usepackage{algorithm}
\usepackage{algorithmic}

\usepackage{graphicx}
\usepackage{tikz}
\usepackage{comment}
\usepackage{amsmath,amssymb} 
\usepackage{color}
\usepackage{booktabs}

\usepackage{afterpage}
\usepackage{colortbl}
\usepackage{makecell}

\usepackage{algorithm}
\usepackage{algorithmic}
\usepackage{amsmath}
\usepackage{multirow}
\newcommand{\myparagraph}[1]{\vspace{0.1em}\noindent\textbf{#1}}

\usepackage{graphicx}
\usepackage{tikz}
\usepackage{comment}
\usepackage{amsmath,amssymb} 
\usepackage{color}
\usepackage{booktabs}

\usepackage{afterpage}
\usepackage{colortbl}
\usepackage{makecell}

\usepackage{newfloat}
\usepackage{listings}
\DeclareCaptionStyle{ruled}{labelfont=normalfont,labelsep=colon,strut=off} 
\floatstyle{ruled}
\newfloat{listing}{tb}{lst}{}
\floatname{listing}{Listing}
\title{HybridCap: Inertia-aid Monocular Capture of Challenging Human Motions}
\author{
    Han Liang\textsuperscript{\rm 1},
    Yannan He\textsuperscript{\rm 1},
    Chengfeng Zhao\textsuperscript{\rm 1},
    Mutian Li\textsuperscript{\rm 1},
    Jingya Wang\textsuperscript{\rm 1}\textsuperscript{\rm 2},\\
    Jingyi Yu\textsuperscript{\rm 1}\textsuperscript{\rm 2},
    Lan Xu\textsuperscript{\rm 1}\textsuperscript{\rm 2}\thanks{Corresponding author}
}
\affiliations{
    \textsuperscript{\rm 1}School of Information Science and Technology, ShanghaiTech University\\
    \textsuperscript{\rm 2}Shanghai Frontiers Science Center of Human-centered Artificial Intelligence\\

    \{lianghan, heyn, zhaochf, limt1, wangjingya, yujingyi, xulan1\}@shanghaitech.edu.cn
}

\begin{document}

\maketitle

\begin{abstract}
Monocular 3D motion capture (mocap) is beneficial to many applications. The use of a single camera, however, often fails to handle occlusions of different body parts and hence it is limited to capture relatively simple movements. 
We present a light-weight, hybrid mocap technique called HybridCap that augments the camera with only 4 Inertial Measurement Units (IMUs) in a learning-and-optimization framework. 
We first employ a weakly-supervised and hierarchical motion inference module based on cooperative pure residual recurrent blocks that serve as limb, body and root trackers as well as an inverse kinematics solver. 
Our network effectively narrows the search space of plausible motions via coarse-to-fine pose estimation and manages to tackle challenging movements with high efficiency. 
We further develop a hybrid optimization scheme that combines inertial feedback and visual cues to improve tracking accuracy. 
Extensive experiments on various datasets demonstrate HybridCap can robustly handle challenging movements ranging from fitness actions to Latin dance.
It also achieves real-time performance up to 60 fps with state-of-the-art accuracy.
\end{abstract}

\section{Introduction}
The past ten years have witnessed a rapid development of human motion capture~\cite{Davison2001,hasler2009markerless,StollHGST2011,wang2017outdoor}, which benefits broad applications like VR/AR, gaming, sports and movies.
However, capturing challenging human motions in a light-weight and convenient manner remains unsolved.

\begin{figure}[tbp] 
	\centering  
	\includegraphics[width=1.0\linewidth]{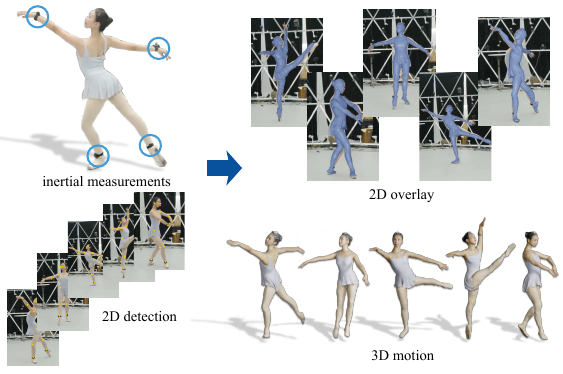} 
	\caption{Our HybridCap achieves robust 3D capture of challenging human motions from a single RGB camera and only 4 IMUs.} 
	\label{fig:fig_1_teaser} 
	\vspace{-9pt} 
\end{figure}

The high-end vison-based solutions require attaching dense optical markers~\cite{VICON} or expensive multi-camera setup~\cite{StollHGST2011,joo2015panoptic,collet2015high,TotalCapture} to capture professional motions, which are undesirable for consumer-level usage.

Recent learning-based methods enable robust human capture from monocular RGB video~\cite{Kanazawa_2019CVPR,VIBE_CVPR2020,DeepHuman_2019ICCV,Xiang_2019_CVPR}.
They require specific human templates for space-time coherent capture~\cite{LiveCap2019tog,MonoPerfCap,EventCap_CVPR2020,DeepCap_CVPR2020}, or utilize parametric human model~\cite{HMR18,AMASS_ICCV2019,SPIN_ICCV2019,VIBE_CVPR2020,SMPL2015}.
However, these monocular methods are fragile to capture specific challenging motions such as fitness actions or Latin dance, which suffer from complex motion patterns and severe self-occlusion. 
Recent advances compensate occlusions and challenging motions using probabilistic or attention-based partial occlusion modeling~\cite{PHMR_ICCV2021,PARE_ICCV2021}, or using the generative or weakly-supervised prior~\cite{HUMOR_ICCV2021,challencap}.
But the above methods still suffer from the inherent monocular ambiguity due to the lack of reliable observation for self-occluded regions.

In stark contrast, combining occlusion-unaware body-worn sensors for robust motion capture has been widely explored~\cite{henschel2020accurate,pons2011outdoor,gilbert2019fusing,zhang2020fusing,kaufmann2021pose}. 
Utilizing Inertial Measurement Units (IMUs) for motion inertia recording is a very popular trend.
However, most previous works~\cite{pons2010multisensor,pons2011outdoor,von2016human,malleson2017real,vonMarcard2018,gilbert2019fusing,malleson2019real,zhang2020fusing} utilize multi-view video or a relatively large amount of IMUs (from 8 to 17), which is undesirable for daily usage.
%
Recently, learning-based approaches~\cite{huang2018DIP,TransPose2021} enable a real-time motion capture with only 6 IMUs. 
But the lack of visual cue leads to inherent drifts and overlay artifacts. 
In this paper, we tackle the above challenges and present \textit{HybridCap} -- a high-quality inertia-aid monocular approach for capturing challenging human motions, as shown in Fig.~\ref{fig:fig_1_teaser}.
We revisit the light-weight hybrid setting using a single RGB camera and 
sparse IMUs (only 4 in our setting), and reframe such hybrid motion capture into the neural data-driven realm.
It enables practical and robust human motion capture under challenging scenarios at 60 fps with state-of-the-art accuracy.

To this end, we introduce a learning-and-optimization framework, consisting of a hybrid motion inference module and a robust hybrid optimization scheme.
The former is formulated in a weakly supervised and hierarchical multi-stage framework.
Specifically, we adopt a fully differentiable architecture in an analysis-by-synthesis fashion without explicit 3D ground truth annotation.
Thus, for dense hybrid weak supervision, we propose a new largest multi-modal human motion dataset called Hybrid Challenging Motions (HCM) consisting of 176-minute challenging motions recorded from 14 RGB cameras and 17 IMUs with a total of 8.9M images.

Our network takes the sequential 2D pose estimation and motion inertia from our light-weight hybrid setting as input, and learns to infer 3D human motion by comparing predicted tracking results against the dense hybrid observations.
To handle the challenging motions, we further formulate our network into a series of cooperative subtasks, including a limb tracker, a body tracker, a root tracker and a hybrid IK (inverse kinematics) solver. 
For all these four subtasks, we also propose a novel bone tracking block design consisting of light-weight pure sequential recurrent layers with skip connection for temporal motion state modeling, which is trained to consider bone length while tracking or solving.
Such hierarchical multi-stage design makes full use of our hybrid weak supervision by %
progressively narrowing the search space of plausible motions in a coarse-to-fine manner, so as to enable efficient and effective inference of challenging motions.  

Finally, besides the data-driven motion characteristics from previous module, current multi-modal input also encodes reliable visual and inertial hints, especially for the non-occluded regions.
Thus, we further propose a robust hybrid optimization scheme with inferential prior to 
refine the inferred results.
It directly fits 2D and inertial observations to improve the accuracy while keeping multi-modal inferential characteristics to preserve motion plausibility.

To summarize, our main contributions include: 
\begin{itemize} 
\setlength\itemsep{0em}
	\item A real-time and accurate motion capture approach
	augmenting the light-weight monocular setting using only 4 IMUs aiding, achieving significant superiority to state-of-the-arts.

	\item A novel network with a hierarchical framework containing cooperative bone tracking blocks with bone length awareness, and a robust optimization scheme combining inferential prior with input observations, improving capture results effectively.
	
	\item A new largest multi-modal human motion dataset containing a wide range of challenging motions along with abundant records of RGB cameras and IMUs.

\end{itemize}

\section{Related Work}

Based on our inertia-aid monocular setup, we focus on human motion capture from optical and inertial solutions.

\myparagraph{Optical Motion Capture.}
Marker-based methods~\cite{VICON,Vlasic2007} have achieved success in capturing professional human motions which are widely utilized in industry, but they are inapplicable for daily usage due to expensive and tedious setup. 
The exploration of markerless mocap~\cite{BreglM1998,AguiaSTAST2008,TheobASST2010} has made great progress in order to get rid of body-worn markers and pursue efficiency.
Benefiting from researches on parametric human models~\cite{SCAPE2005,SMPL2015,SMPLX2019,STAR_ECCV2020} in the last decade, various data-driven approaches are proposed to estimate 3D human pose and shape by optimizing~\cite{TAM_3DV2017,Lassner17,keepitSMPL,Kolotouros_2019_CVPR} or directly regressing~\cite{HMR18,Kanazawa_2019CVPR,VIBE_CVPR2020,zanfir2020neural} human model parameters. 
Taking specific template mesh as prior, multi-view~\cite{Gall2010,StollHGST2011,liu2013markerless,Robertini:2016,Pavlakos17,Simon17,FlyCap} and monocular~\cite{MonoPerfCap,LiveCap2019tog,EventCap_CVPR2020,DeepCap_CVPR2020} template-based approaches combine free-form and parametric methods, which produce high quality skeletal and surface motions.
Besides, to alleviate the inherent estimation ambiguity from 2D input to 3D motion, recent approaches~\cite{PHMR_ICCV2021,PARE_ICCV2021} handle complex patterns using probabilistic or attention-based semantic modeling, \cite{HUMOR_ICCV2021,challencap} learn to model generative or weakly-supervised prior to solve unseen and non-periodic motions. 
However, these methods still suffer from challenging motions, especially for rare pose patterns and extreme self-occlusion.

\begin{figure*}[t]
	\centering
	\includegraphics[width=\linewidth]{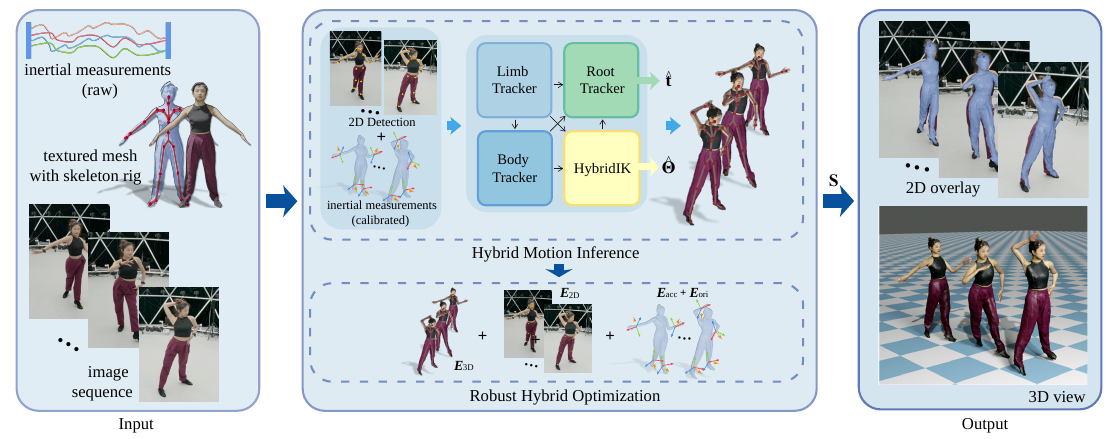}
	\caption{The pipeline of HybridCap with multi-modal input. Our approach combines a hybrid motion inference stage with a robust hybrid motion optimization stage to capture 3D challenging motions.  } 
	\label{fig:overview}
	\vspace{-3mm}
\end{figure*}

\myparagraph{Inertial Motion Capture.}
To overcome the limitations of vision cues only, another category of works propose to use IMUs. 
Previously, purely inertial methods using large amounts of sensors like Xsens MVN~\cite{XSENS} has been commercially used. However, intrusive capture system prompts researchers forward to sparse-sensor setup. SIP~\cite{von2017SIP}, which uses only 6 IMUs, presents a pioneering exploration. However, limitations of its traditional optimization framework make real-time application impractical. Recent data-driven works~\cite{huang2018DIP,TransPose2021,PIPCVPR2022} achieve great improvements on accuracy and efficiency with sparse sensors, but substantial drift is still unsolved for challenging motions.
Preceding sensor-aid solutions propose to combine IMUs with videos~\cite{gilbert2019fusing,henschel2020accurate,malleson2019real,malleson2017real}, RGB-D cameras~\cite{helten2013real,Zheng2018HybridFusion}, optical markers~\cite{Andrews2016} 
or even LiDAR~\cite{10049734}. 
Although these approaches partially solve scene-occlusion problem and correct drift effectively, they are restricted from either undesirable system complexity or implausible estimation for challenging motions.

\begin{figure*}[t]
	\centering
	\includegraphics[width=\linewidth]{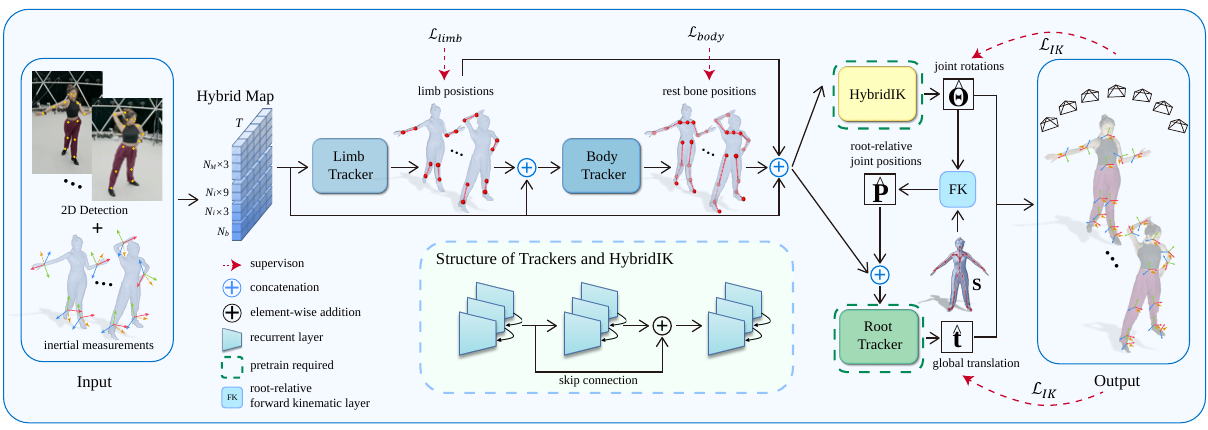}
	\caption{Illustration of our hybrid motion inference module, which is based on cooperative pure residual recurrent blocks that serve as limb, body and root trackers as well as a hybrid inverse kinematics (IK) solver. The limb tracker focuses on accurately tracking for 4 limbs while the body tracker estimates the rest bone positions from accurate limb positions. Then, the hybrid IK solver and root tracker are employed to combine initial input with well-estimated bone positions to solve rotations $\mathbf{\hat{\Theta}}$ and global translation $\mathbf{\hat{t}}$.}
	\label{fig:fig_3_network}
	\vspace{-3mm}
\end{figure*}

\section{Preliminary and Overview}
The goal of our work is to capture challenging 3D human motions using a single camera with few inertial sensors aiding, which suffers from complex motion patterns and extreme self-occlusion.
Fig.~\ref{fig:overview} provides an overview of HybridCap, which relies on a template mesh of the actor and makes full usage of multi-modal input in a learning-and-optimization framework.
In the inference stage, our hierarchical design extracts different characteristics from multi-modal observations and learns to estimate plausible motions in a weakly supervised manner. 
Then, a robust optimization stage is introduced to refine the skeletal motions to increase the tracking accuracy and overlay performance.

\myparagraph{Template and Motion Representation.}
We first scan the actor with a 3D body scanner to generate the textured template mesh of the actor. Then, we rig it automatically by fitting the Skinned Multi-Person Linear Model (SMPL)~\cite{SMPL2015} to the template mesh and transferring the SMPL skinning weights to our scanned mesh.
The kinematic skeleton is parameterized as $\textbf{S}=[\mathbf{\theta}, \textbf{R},\textbf{t}]$, including the Euler angles $\mathbf{\theta} \in \mathbb{R}^{N_J\times3}$ of the $N_J$ joints, the global rotation $\textbf{R}\in\mathbb{R}^3$ and translation $\textbf{t} \in \mathbb{R}^{3}$. 
Furthermore, let $\mathbf{\Theta}$ denotes the 6D representation \cite{zhou2019continuity} of the global rotation and joint rotations.
Then, we can formulate $\textbf{S}=\mathcal{M}(\mathbf{\Theta},\textbf{t})$ where $\mathcal{M}$ denotes the motion transformation between various representations.

\myparagraph{Input Preprocessing.}
Our system takes the RGB video, inertial measurements and a well pre-scanned template of the actor as the overall input.
Given an image frame, we extract $N_M$ 2D keypoints $\boldsymbol{p} \in \mathbb{R}^{N_M\times2}$ and corresponding confidence $\boldsymbol{\sigma} \in \mathbb{R}^{N_M}$ using OpenPose \cite{OpenPose}. To generalize to various camera settings during inference time, we refer \cite{PhysAwareTOG2021} to use canonicalized 2D keypoints $\boldsymbol{p}_c$ by projecting them onto $Z=1$ plane. 
Then we transform inertial measurements from inertial frame $\mathcal{F}_I$ into camera frame $\mathcal{F}_C$ and obtain IMU accelerations $\mathbf{A}_{n}\in \mathbb{R}^{3}$ and orientations $\mathbf{R}_{\textbf{b}_n}\in \mathbb{R}^{3\times3}$ of $N_i$ corresponding bones $\textbf{b}_n$ with calibrated $\mathbf{R}_{I2C}$ and $\mathbf{R}_{S2B,n}$:
\begin{align}
    &\mathbf{A}_{n} = \mathbf{R}_{I2C} \mathbf{A}_{I,n}\\
    &\mathbf{R}_{\textbf{b}_n} = \mathbf{R}_{I2C} \mathbf{\widetilde{R}}_n \mathbf{R}_{S2B, n}^T,
\end{align} 
where $\mathbf{R}_{I2C}$ is the transformation from $\mathcal{F}_I$ to $\mathcal{F}_C$, and $\mathbf{R}_{S2B,n}$ is the transformation from the $n$-th IMU sensor $\mathcal{F}_{\textbf{s}_n}$ to $\mathcal{F}_{\textbf{b}_n}$ of its corresponding bone $\textbf{b}_n$.
Besides, to provide prior knowledge on anthropometry, we heuristically calculate $N_b=7$ key bone lengths $\mathbf{L_k}$ (uparm, lowarm, upleg, lowleg, foot, clavicle, and spine) from the rigging skeleton $\mathbf{S}$ and concatenate them into the input. Thus the overall input of the network in a single frame is $[\boldsymbol{p}_c, \boldsymbol{\sigma}, \mathbf{R}_{\textbf{b}}, \mathbf{A},  \mathbf{L_k}]\in \mathbb{R}^{2N_M+N_M+9N_i+3N_i+N_b}$.

\begin{figure*}[th]
	\centering
	\includegraphics[width=\linewidth]{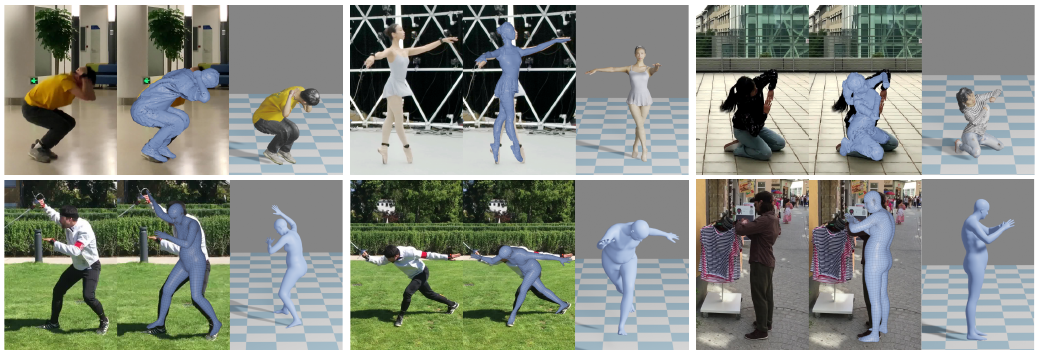}
	\caption{Our qualitative results. Each picture shows the input image, 2D overlay and motion in 3D space from left to right respectively. The results show that our approach produces good 2D overlay and plausible 3D motion.}
	\label{fig:gallery}
	\vspace{-1mm}
\end{figure*}

\section{Approach}
\subsection{Hybrid Motion Inference} \label{sec:mot_inference}
As illustrated in Fig.~\ref{fig:fig_3_network}, we adopt a hierarchical multi-stage motion inference scheme.
It takes the sequential 2D pose estimation and motion inertia as input and predicts 3D human motions in a weakly supervised manner without explicit 3D ground truth annotation.
To tackle challenging motions, we divide the motion inference task into a series of cooperative subtasks.
Specifically, we introduce a limb tracker, a body tracker, a root tracker, and a hybrid IK (inverse kinematics) solver, so as to model the hierarchical knowledge of articulated human body structure.

\myparagraph{Pure Residual Recurrent Blocks.}
For temporal motion state modeling in these subtasks, we propose a novel efficient block design using pure recurrent layers with skip connection.
In contrast to block design in TransPose~\cite{TransPose2021} adopting both recurrent (LSTM) and fully-connected (FC) layers, we find that such FC layers are the key bottleneck, which encode non-temporal representation and tend to overfit. 
Thus we 
simply remove the FC layers and improve the accuracy.

We further introduce skip connection. Our key intuition is that motion inference or position tracking requires less high-level features. 
We adopt skip connection enabling easier identity function learning to retain more information from the input layer, where the layer close to the output is responsible for adding low-level details.
Thus the low-level 2D and inertial features could be selected and passed to the output layer directly, which helps more accurate and detailed motion inference.

\myparagraph{Hierarchical Bone Trackers.}
In our hierarchical design, our trackers track bones rather than joints. For a bone $\textbf{b}_n$, we track root-relative positions of its two endpoint $\mathbf{\hat{J}}$ with distance constraint using the known bone lengths $\mathbf{L}_{\textbf{b}_n}$. 
Specifically, the limb tracker focuses on accurately tracking four limbs with reliable inertial measurements. Here, the loss of limb tracker is formulated as:
\begin{align}
	\mathcal{L}_{limb} = \mathcal{L}_{joint}^{(limb)} + \mathcal{L}_{bone}^{(limb)},
\end{align}
where $\mathcal{L}_{joint}^{(limb)}$ is the 3D joint position loss and $\mathcal{L}_{bone}^{(limb)}$ is the limb bone length loss. 
Next, the body tracker estimates the rest rigid body parts concatenating the initial input and well-estimated limb positions as the input. Bone length constraints are also utilized to reduce depth ambiguity. Furthermore, the bone orientations help to narrow the search space to the target joint positions. Similar to the limb tracker loss, the body tracker loss is formulated as:
\begin{align}
	\mathcal{L}_{body} = \mathcal{L}_{joint}^{(body)} + \mathcal{L}_{bone}^{(body)}.
\end{align}
Note that $\mathcal{L}_{joint}$ is formulated as the reprojection error to all camera views, which guides the corresponding tracker to predict joint positions in a weakly supervised manner:
\begin{align}
	\mathcal{L}_{joint} = \sum_{t=1}^T \sum_{c=1}^{N_C} \sum_{j=1}^{N_J} \sigma_{c,j}^{(t)}||\mathbf{\Pi}_c(\mathbf{\hat{J}}_j^{(t)}+\textbf{t})  - \mathbf{p}_{c,j}^{(t)}||_2^2,
\end{align}
where $\sigma_{c,j}^{(t)}$ denotes the confidence of 2D joint $\mathbf{p}_{c,j}^{(t)}$; $\mathbf{\Pi}_c$ denotes the projection function of camera $c$; $\mathbf{t}$ denotes global translation from hybrid optimization of full observations; $N_J$ denotes the number of bone endpoints (8 for limb tracker and 7 for body tracker). 
Then, the bone length loss is formulated as the $L_2$ loss between the predicted bone length and ground-truth bone length $\mathbf{L}_{\textbf{b}_n}$:
\begin{align}
	\mathcal{L}_{bone} = \sum_{t=1}^T \sum_{n=1}^{N_B} (||\mathbf{\hat{J}}_{\textbf{b}_n, 0}^{(t)} - \mathbf{\hat{J}}_{\textbf{b}_n, 1}^{(t)}||_2 - \mathbf{L}_{\textbf{b}_n})^2,
\end{align}
%
where the predicted bone length can be calculated by the distance of two output endpoint positions $\mathbf{\hat{J}}_{\textbf{b}_n, i}^{(t)} (i=0,1)$ of bone $\textbf{b}_n$. Note that $N_B$ is the number of target bones which is 4 for the limb tracker and 8 for the body tracker.

\myparagraph{Hybrid Inverse Kinematics Solver.}
Based on the accurate tracking of bones, we introduce our hybrid IK solver and root tracker to solve rotations and translation respectively. The initial input and well-estimated root-relative 3D joints are concatenated and fed into our hybrid IK solver, which outputs global rotation and local joint rotations $\mathbf{\hat{\Theta}}$ in the 6D representation. Then we perform forward kinematics using predicted rotations $\mathbf{\hat{\Theta}}$ and bone lengths to obtain refined root-relative 3D joints. Next we send them with the initial input into the root tracker which predicts root position $\hat{\textbf{t}}$ (i.e. global translation) in the camera frame. 

To utilize our dense hybrid weak supervision, we
further
calculate $N_M$ 3D marker positions $\mathbf{\hat{P}}_m(\mathbf{\hat{\Theta}}, \hat{\textbf{t}})$ attached to the skeleton (corresponding to body joints and face landmarks of OpenPose), $N_I$ bone orientations $\mathbf{\hat{R}}_{\textbf{b}_n}(\mathbf{\hat{\Theta}})$, 
and $N_I$ simulated IMU sensor positions $\mathbf{\hat{P}}_n(\mathbf{\hat{\Theta}}, \hat{\textbf{t}})$ 
respectively. The IK loss
is formulated as:
\begin{align}
	\mathcal{L}_{IK} = \lambda_{\mathrm{2D}}\mathcal{L}_{2D} + \lambda_{\mathrm{acc}}\mathcal{L}_{acc}+ \lambda_{\mathrm{ori}}\mathcal{L}_{ori} \nonumber\\
	+ \lambda_{\mathrm{prior}}\mathcal{L}_{prior} + \lambda_{\mathrm{trans}}\mathcal{L}_{trans}.
\end{align}
The 2D reprojection loss $\mathcal{L}_{2D}$ ensures each estimated 3D marker $\mathbf{\hat{P}}_{m}$ projects onto the corresponding 2D keypoint $\mathbf{p}_{c,m}$ in all camera views, formulated as:
\begin{align}
	\mathcal{L}_{2D} = \sum_{t=1}^T \sum_{c=1}^{N_C} \sum_{m=1}^{N_M} \sigma_{c,m}^{(t)}||\mathbf{\Pi}_c(\mathbf{\hat{P}}_{m}^{(t)}(\mathbf{\hat{\Theta}}, \hat{\textbf{t}}))  - \mathbf{p}_{c,m}^{(t)}||_2^2,
\end{align}
where $\sigma_{c,m}^{(t)}$ denotes the confidence of 2D keypoint $\mathbf{p}_{c,m}$ and $\mathbf{\Pi}_c$ is the projection function of camera $c$.
Then, we introduce the acceleration loss $\mathcal{L}_{acc}$ to encourage the network to learn the implicit physical constraints and generate plausible motions:
\begin{align}
	\mathcal{L}_{acc} = \sum_{t=2}^{T-1} \sum_{n=1}^{N_I} ||\mathbf{\hat{A}}_n^{(t)}(\mathbf{\hat{\Theta}}, \hat{\textbf{t}})  - \mathbf{A}_n^{(t)}||_2^2,
\end{align}
where $\mathbf{\hat{A}}_n^{(t)}$ is the estimated acceleration calculated from predicted IMU position $\mathbf{\hat{P}}_n$, formulated as below where $st$ is the sampling time:
\begin{equation}
	\mathbf{\hat{A}}_n^{(t)} = (\mathbf{\hat{P}}_n^{(t+1)} - 2\mathbf{\hat{P}}_n^{(t)} + \mathbf{\hat{P}}_n^{(t-1)}) / st^2.
	\label{eq:acceleration}
\end{equation}
We further propose the orientation loss $\mathcal{L}_{ori}$, which ensures the predicted orientation $\mathbf{\hat{R}}_{\textbf{b}_n}$ of each bone bound with IMU sensor fits the observation $\mathbf{R}_{\textbf{b}_n}$ of the corresponding IMU measurement.
\begin{align}
	\mathcal{L}_{ori} = \sum_{t=1}^T \sum_{n=1}^{N_I} ||\mathbf{\hat{R}}_{\textbf{b}_n}^{(t)}(\mathbf{\hat{\Theta}})  - \mathbf{R}_{\textbf{b}_n}^{(t)}||_2^2.
\end{align}
Furthermore, the prior loss $\mathcal{L}_{prior}$ is the $L_2$ loss between predicted rotations $\mathbf{\hat{\Theta}}$ and reference $\mathbf{\Theta}$, while $\mathcal{L}_{trans}$ the one related to the predicted root position $\hat{\textbf{t}}$ and reference position $\textbf{t}$.
Both $\mathbf{\Theta}$ and $\textbf{t}$ are dense ``pseudo ground-truth'' references.

\begin{table*}[th]
	\begin{center}
		\centering
		
 		\resizebox{1.0\textwidth}{!}{
		\begin{tabular}{l|ccc|ccc|ccc}
		    \hline
		    \multirow{2}{*}{Method}  & \multicolumn{3}{c|}{HCM}& \multicolumn{3}{c|}{3DPW}& \multicolumn{3}{c}{AIST++}\\
			& MPJPE$\downarrow$ & PCK@0.2$\uparrow$  & Accel error$\downarrow$& MPJPE $\downarrow$ & PCK@0.2$\uparrow$  & Accel error$\downarrow$& MPJPE $\downarrow$ & PCK@0.2$\uparrow$  & Accel\\
			\hline
			VIBE         & 103.4  & 61.5   & 96.8   & 82.9  & 70.3 & 21.1 &  73.5  &  69.1  & 94.6 \\
			HuMoR        & 81.9  &  72.7  &  31.8  & 77.0  & 75.8 & 13.2   &  59.4 &  81.3  & 33.1\\
			TransPose & 73.1 & 77.4   & 65.6   & 76.5  & 76.9  & 19.3  & 61.6  & 78.2   & 57.0 \\
			ChallenCap & 69.5 &  79.5  &  98.1  & 78.2  &  74.2 & 48.4   &  53.2 &  86.8  & 91.4\\
			\rowcolor{black!7}  Ours        &\textbf{43.3}  & \textbf{90.1}   & \textbf{17.9}   & \textbf{72.1}&  \textbf{80.5}  &   \textbf{5.4}& \textbf{33.3}& \textbf{95.4}& 14.4 \\
			\hline
		\end{tabular}
 		}
        \caption{Quantitative comparison of several previous state-of-the-art methods in terms of tracking accuracy and plausibility. }
		\label{tab:Comparison}
	\end{center}
	\vspace{-3mm}
\end{table*}

\subsection{Hybrid Motion Optimization}
\label{sec:mot_optimization}

Despite the data-driven motion inference stage learns the mapping from multi-modal observations to 3D motions, the generalization error is non-negligible due to noisy 2D detection and inertial measurements. 

We further introduce a hybrid motion optimization stage to refine the skeletal motions to increase the tracking accuracy and overlay performance.
It jointly utilizes the learned 3D prior from the network of multi-modal weak supervision, the 2D keypoints in the visible regions as well as inertial measurements. 

In this stage, we first initialize the skeletal motion sequence $\textbf{S}$ using network output by representation transformation $\mathcal{M}(\mathbf{\hat{\Theta}},\mathbf{\hat{t}})$ and then perform the optimization procedure. 
We adopt the Euler angle representation so that the joint angles $\boldsymbol{\theta}$ of $\textbf{S}$ locate in the pre-defined range $[\boldsymbol{\theta}_{min},  \boldsymbol{\theta}_{max}]$ of physically plausible joint angles to prevent unnatural poses.
Our energy function is formulated as:
\begin{align} \label{eq:opt}
	\vspace{-0.2mm}
	\boldsymbol{E}(\mathbf{S}) = \lambda_{\mathrm{3D}}\boldsymbol{E}_{\mathrm{3D}} + \lambda_{\mathrm{2D}}\boldsymbol{E}_{\mathrm{2D}} +
	\lambda_{\mathrm{acc}}\boldsymbol{E}_{\mathrm{acc}} + \lambda_{\mathrm{ori}}\boldsymbol{E}_{\mathrm{ori}}.
	\vspace{-0.2mm}
\end{align}
Here, $\boldsymbol{E}_{\mathrm{3D}}$ enforces the final motion sequence close to the predicted one; $\boldsymbol{E}_{\mathrm{2D}}$ ensures that each final 3D marker reprojects onto the corresponding 2D keypoint.

Besides, we adopt the acceleration energy $\boldsymbol{E}_{\mathrm{acc}}$ to enforce the final motion to be temporally consistent with the network estimating accelerations $\mathbf{\hat{A}}_n$ supervised by $N_I$ IMUs and the measured ground-truth accelerations $\mathbf{A}_n$ from $N_i$ input IMUs.
Specifically, the acceleration term $\boldsymbol{E}_{\mathrm{acc}}$ is formulated as:
\begin{align}
\label{Eqn.acceleration term}
\boldsymbol{E}_{\mathrm{acc}} =  &\sum_{t=2}^{T-1} \sum_{n=1}^{N_i} \gamma_{n}^{(t)}||\mathbf{A}_n^{(t)}(\mathbf{S})  - \mathbf{A}_n^{(t)}||_2^2 \nonumber\\
&+\sum_{t=2}^{T-1} \sum_{n=N_i+1}^{N_I}||\mathbf{A}_n^{(t)}(\mathbf{S})  - \mathbf{\hat{A}}_n^{(t)}||_2^2,
\end{align}
where the first term means that we directly use the acceleration observation from the input IMUs and the second term implies that we trust the network for those unobserved parts. 
The final acceleration $\mathbf{A}_n^{(t)}(\mathbf{S})$ is obtained from skeletal motion sequence $\mathbf{S}$ as same as Eqn.~\ref{eq:acceleration}.
Then, we adopt the orientation energy $\boldsymbol{E}_{\mathrm{ori}}$ to enforce the final bone orientations to be consistent with the observations of $N_i$ input IMUs, which is formulated as: 
\begin{align} \label{eq:opt_ori}
	\boldsymbol{E}_{\mathrm{ori}} = \sum_{t=1}^T \sum_{n=1}^{N_i} ||\mathbf{R}_{\textbf{b}_n}^{(t)}(\mathbf{\mathbf{S_t}})  - \mathbf{R}_{\textbf{b}_n}^{(t)}||_2^2,
\end{align}
where $\mathbf{R}_{\textbf{b}_n}^{(t)}(\mathbf{\mathbf{S_t}})$ denotes the final orientation of bone $\textbf{b}_n$, which is obtained from skeleton $\mathbf{S_t}$.

\myparagraph{Optimization.}
The constrained optimization problem to minimize the Eqn.~\ref{eq:opt} is solved using the Levenberg-Marquardt (LM) algorithm of ceres~\cite{ceresSolver} 
with 4 iterations. In our experiments, we empirically set the parameters $\lambda_{\mathrm{3D}}=10$, $\lambda_{\mathrm{2D}}=1$, 
$\lambda_{\mathrm{acc}}=10$ and $\lambda_{\mathrm{ori}}=30$.

\begin{figure}[b]
	\centering
	\includegraphics[width=\linewidth]{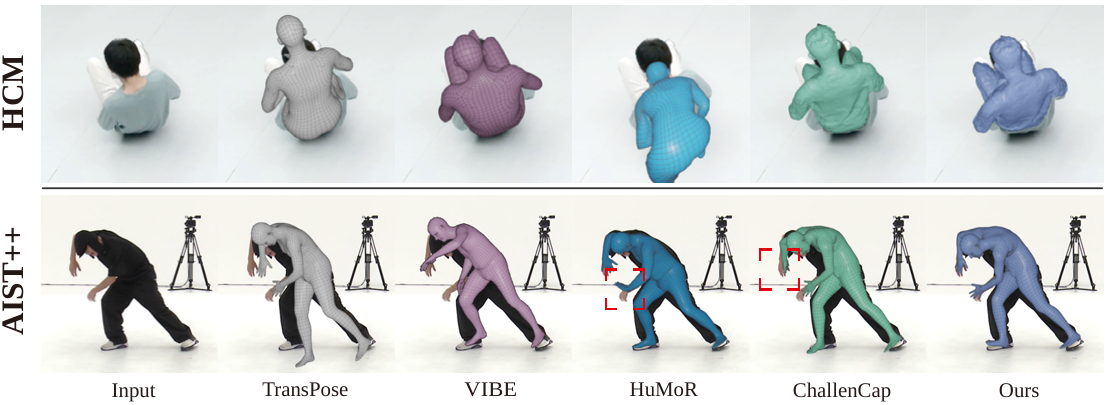}
	\caption{Our method outperforms the previous state-of-the-arts on the performance of both overlay and limb orientation.}
	\label{fig:Comparison}
\end{figure}
\section{Experiments}
\label{ExperimentalResults}
\begin{figure}[t]
	\centering
	\includegraphics[width=1.0\linewidth]{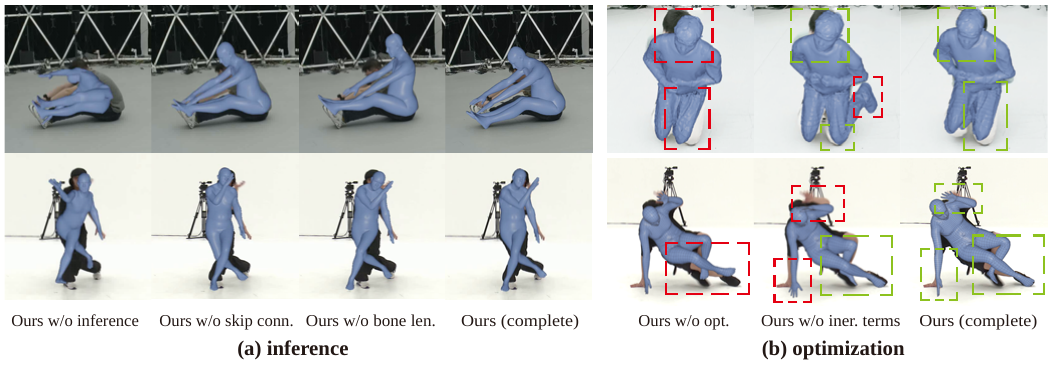}
	\caption{Qualitative evaluation on our (a) inference module and (b) optimization module. }
	\label{fig:ablation_net}
\end{figure}

\begin{figure}[t]
	\centering
	\includegraphics[width=1.0\linewidth]{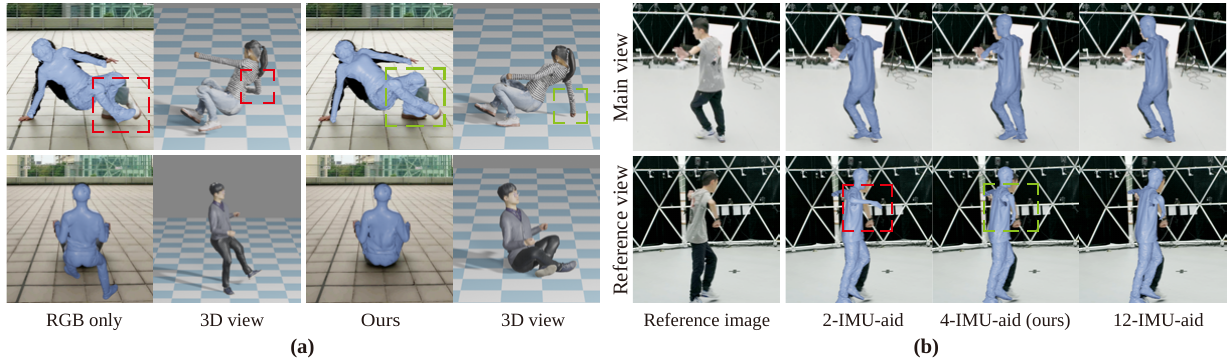}
	\caption{Qualitative evaluation on our (a) multi-modal input setting and (b) IMU number configurations.}
	\label{fig:ablation_optimize}
\end{figure}

We run our pipeline on a PC with an i7-10700k CPU and RTX3070 GPU, where the inference module and optimization module take 2.8($\pm$0.4) ms and 6.2($\pm$3.5) ms respectively, achieving 60 fps. 
In this section, first, we describe the datasets used for training and evaluation. 
Next, we further qualitatively (Fig.~\ref{fig:Comparison}) and quantitatively (Tab.~\ref{tab:Comparison}) illustrate that our method outperforms previous state-of-the-arts.
We also provide extensive ablation studies (Tab.~\ref{tab:ablation_framework}, Tab.~\ref{tab:ablation_input} and Fig.~\ref{fig:ablation_net}, Fig.~\ref{fig:ablation_optimize}) to evaluate our technical design and input setting.
Finally, we present more qualitative results in Fig.~\ref{fig:gallery}.

\myparagraph{Training Dataset.}
We use mixed multi-modal datasets to train our inference module.
To provide sufficient 2D and inertial observations, we build and propose a new dataset Hybrid Challenging Motions (HCM), which consists of abundant records of RGB cameras and IMUs. Please refer to the supplementary to appreciate more details.
We further utilize AIST++~\cite{li2021learn,aist-dance-db} which consists of various challenging dance sequences. 
Note that when using AIST++ we simulate $N_i$ IMU measurements only as the input rather than supervision. 

\myparagraph{Evaluation Dataset.}
For thorough evaluation, we use 3DPW~\cite{vonMarcard2018}, AIST++, and our HCM, which contain complex in-the-wild scenarios or challenging motions. We follow the standard protocols to split out test sets for all the datasets.
Note that we use 3DPW that only provides monocular data for testing rather than training. 
Our code and HCM dataset will be made publicly available to stimulate future research.

\myparagraph{Metrics.}
To evaluate capture accuracy, we report the Mean Per Joint Position Error (MPJPE) in terms of \textit{mm}, and the Percentage of Correct Keypoints (PCK) with 0.2 torso size as the threshold. For HCM and 3DPW with ground-truth acceleration measured by IMUs, we report acceleration error ($m/s^2$), calculated as the difference between the 
ground-truth and estimated acceleration. For AIST++, we report mean per-joint accelerations (Accel)~\cite{Kanazawa_2019CVPR,HUMOR_ICCV2021} to evaluate motion plausibility.

\vspace{-2mm}
\subsection{Comparison}
\vspace{-1mm}
We compare HybridCap with various representative methods using a single RGB camera or sparse IMUs. 
Specifically, we apply ChallenCap~\cite{challencap} and HuMoR~\cite{HUMOR_ICCV2021} which are also based on a learning-and-optimization framework, VIBE~\cite{VIBE_CVPR2020} using adversarial learning, as well as TransPose~\cite{TransPose2021} using pure 6 IMUs. 
For fair comparisons, we fine-tune them with the same training dataset. 
As shown in Fig.~\ref{fig:Comparison}, our method gets better overlays of the captured body and obtains more accurate limb orientation results.
Tab.~\ref{tab:Comparison} provides the corresponding quantitative comparison results under various metrics and datasets. 

Note that our approach consistently outperforms other approaches on various metrics, which denotes the effectiveness of our method to handle multi-modal input.
Besides, our approach focuses on capturing challenging motions with self-occlusion of a single actor.
However, the 3DPW dataset lacks such self-occlusion scenarios and includes those sequences with incomplete 2D observation and external occlusions due to scene interactions. 
Thus, such domain gaps in terms of occlusion lead to the performance drop-off of our approach when evaluating on 3DPW. 
Nevertheless, we still achieve better performance.

\begin{table}[t]
	\begin{center}
		\resizebox{0.47\textwidth}{!}{
    		\begin{tabular}{clcccc}
    			\hline
    			     Module & \multicolumn{1}{l}{Ablation setting}  & MPJPE$\downarrow$         & PCK@0.2$\uparrow$         & Accel error $\downarrow$         \\
    			\hline
    			\multirow{4}{*}{Stage 1} & 1) VIBE structure   & 75.4        & 76.5              & 34.2          \\
    			& 2) TransPose structure       & 61.0         & 82.7      & 19.2          \\
    			& 3) Ours w/o inference            & 136.5         & 51.1       & 82.1 \\
    			& 4) Ours w/o skip connection            & 54.5    & 86.9            & 18.7         \\
    			& 5) Ours w/o bone length            & 65.8     & 81.3             & 19.6         \\
    			\midrule
    		    \multirow{2}{*}{Stage 2}	& 6) Ours w/o optimization            & 57.5         & 85.3         & 18.6          \\
    			& 7) Ours w/o inertial terms   & 51.2         & 87.4         & 62.2          \\
    
    			\Xhline{0.1\arrayrulewidth}
    			\rowcolor{black!7}  & 8) Ours (complete)         & \textbf{43.3} & \textbf{90.1}  & \textbf{17.9} \\
    			\hline
    		\end{tabular}
		}
		\caption{Quantitative evaluation on our network structure, bone length utilization and optimization configurations. }
	    \label{tab:ablation_framework}
	    \vspace{-4mm}
	\end{center}
\end{table}

\begin{table}[t]
	\begin{center}
		\centering
		\resizebox{0.47\textwidth}{!}{
		\begin{tabular}{llcccc}
			\hline
			Modality& \multicolumn{1}{l}{Input setting}   & MPJPE$\downarrow$         & PCK@0.2$\uparrow$       & Accel error$\downarrow$         \\
			\hline
			\multirow{2}{*}{Single}& 4 IMUs Only   & 104.7         & 57.4      & 18.5          \\
			& RGB Only   & 77.6         & 74.2        & 28.1          \\
			\hline
			
			\multirow{3}{*}{Hybrid}& 2-IMU-aid &      65.0      & 81.0       & 23.4          \\
			& 4-IMU-aid (Ours)&   \textbf{43.3}          & \textbf{90.1}       & \textbf{17.9}          \\
			& 12-IMU-aid & 36.1          & 93.2      & 15.1          \\
			\hline
		\end{tabular}
		}
		\caption{Quantitative evaluation on our input configuration.
		}
		\label{tab:ablation_input}
	\end{center}
	\vspace{-4mm}
\end{table}

\vspace{-2mm}
\subsection{Ablation Study} \label{paper:Evaluation}
\vspace{-1mm}

\myparagraph{Evaluation on inference module.}
We first evaluate our inference module by comparing to the variants of our approach and previous state-of-the-arts. The quantitative results are provided in Rows 1-5 in Tab.~\ref{tab:ablation_framework}. \textbf{VIBE structure} (Row 1) shows VIBE variant with 4 IMUs and RGB input, where we concatenate inertial input and key bone lengths with image features obtained from pre-trained feature extractor used in the vanilla VIBE. \textbf{TransPose structure} (Row 2) shows TransPose variant with 4 IMUs and RGB input, where we concatenate 2D keypoints and key bone lengths with inertial input. We train both two networks using the same training set and losses as ours.  
The results show our network with nuanced designs to progressively fuse multi-modal input is superior to both two previous state-of-the-arts. 
Row 3-5 corresponds to the variants without our entire inference module, without using the skip connection in our block design, without using bone length, respectively. 
\textbf{Purely recurrent design} (Row 4) without FC layers (adopted in Row 2), effectively alleviates overfitting. \textbf{Skip connection} (Row 4) operation enables to preserve more low-level features from the input layer, and the \textbf{bone length} (Row 5) awareness effectively narrows the search space, which are the two key designs to improve the inference results. Corresponding qualitative results are provided in Fig.~6 (a), which demonstrates our two nuanced designs also contribute to improving the overlay performance. 
These results illustrate the effectiveness of inference module and highlight the contribution of our algorithmic component designs.

\myparagraph{Evaluation on optimization module.}
To evaluate the effectiveness of our optimization module design, we further compare to the two variants without the entire optimization module and without using the inertial terms in Eqn.~\ref{Eqn.acceleration term} and Eqn.~\ref{eq:opt_ori}, respectively. The quantitative results are provided in Rows 6-7 of Tab.~\ref{tab:ablation_framework}. Our inertial terms improve position accuracy while preserving motion plausibility (acceleration) from the inference module. The qualitative results provided in Fig.~6 (b) show our robust optimization improves the overlay performance effectively. Note that these variants suffer from tracking loss especially for the limbs with fast and challenging motions (generalization error). In contrast, our full pipeline achieves more robust capture.

\myparagraph{Evaluation on 4-IMU-aid setting.}
Finally, to verify the necessity and rationality of our multi-modal input setting with only 4 IMUs aiding, we compare several variants of our approach with various network input settings. As shown in Tab.~\ref{tab:ablation_input}, even using 2 IMUs aiding, our multi-modal input outperforms pure IMU or RGB input under single modality. Besides, our approach with 4 IMUs significantly outperforms the one with only two IMUs (one for left arm and one for right leg) and closes to the one with tedious 12 IMUs, which serves as a good compromise of acceptable performance and light-weight convenient capture setting.
As shown in Fig.~7 (a), our inertia-aid setting augments the camera by alleviating its inherent defects in terms of depth ambiguity and occlusion. As shown in Fig.~\ref{fig:ablation_optimize} (b), 4-IMU-aid setting enables to constantly track frequently occluded extremities while minimizing the intrusive body-worn sensors.

\section{Conclusion}
We present a practical inertia-aid monocular approach to capture challenging 3D human motions, augmenting the single camera with only 4 IMUs, achieving significant superior to previous state-of-the-arts.   
It implements such an extremely light-weight hybrid capture setting with a learning-and-optimization framework through novel cooperative bone tracking blocks with bone length awareness and an optimization-based refinement module with inferential prior. 
Besides, we propose a new largest multi-modal human motion dataset, called HCM dataset, to evaluate our approach thoroughly.
Our experimental results demonstrate the robustness of HybridCap in capturing challenging human motions in various scenarios.
We believe that it is a significant step for convenient and robust capture of human motions, with many potential applications in VR/AR and motion evaluations for gymnastics and dancing.

\section{Acknowledgements}
This work was supported by Shanghai YangFan Program (21YF1429500), Shanghai Local college capacity building program (22010502800), NSFC programs (61976138, 61977047), the National Key Research and Development Program (2018YFB2100500), STCSM (2015F0203-000-06), and SHMEC (2019-01-07-00-01-E00003).

%
%
\bibliography{aaai23.bib}

\vspace{-2mm}
\section{Appendix}
\begin{figure}[h]
	\centering
	\includegraphics[width=0.8\linewidth]{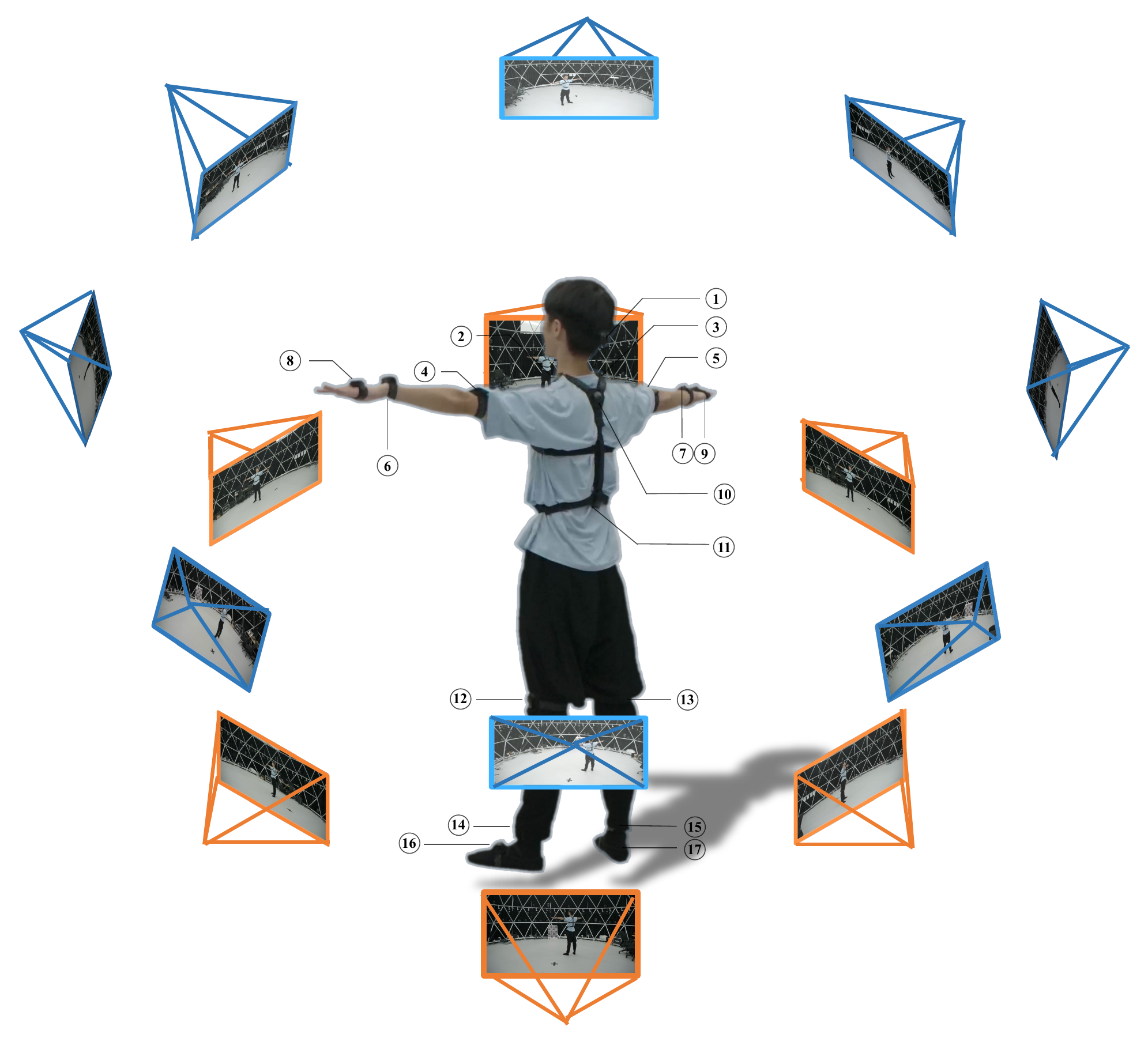}
	\caption{Our HCM dataset collection studio which contains 14 cameras covering a wide range of viewpoints and 17 IMUs covering most human rigid body parts.}
	\label{fig:wear}
\end{figure}

\begin{figure*}[th]
	\centering
	\includegraphics[width=1\linewidth]{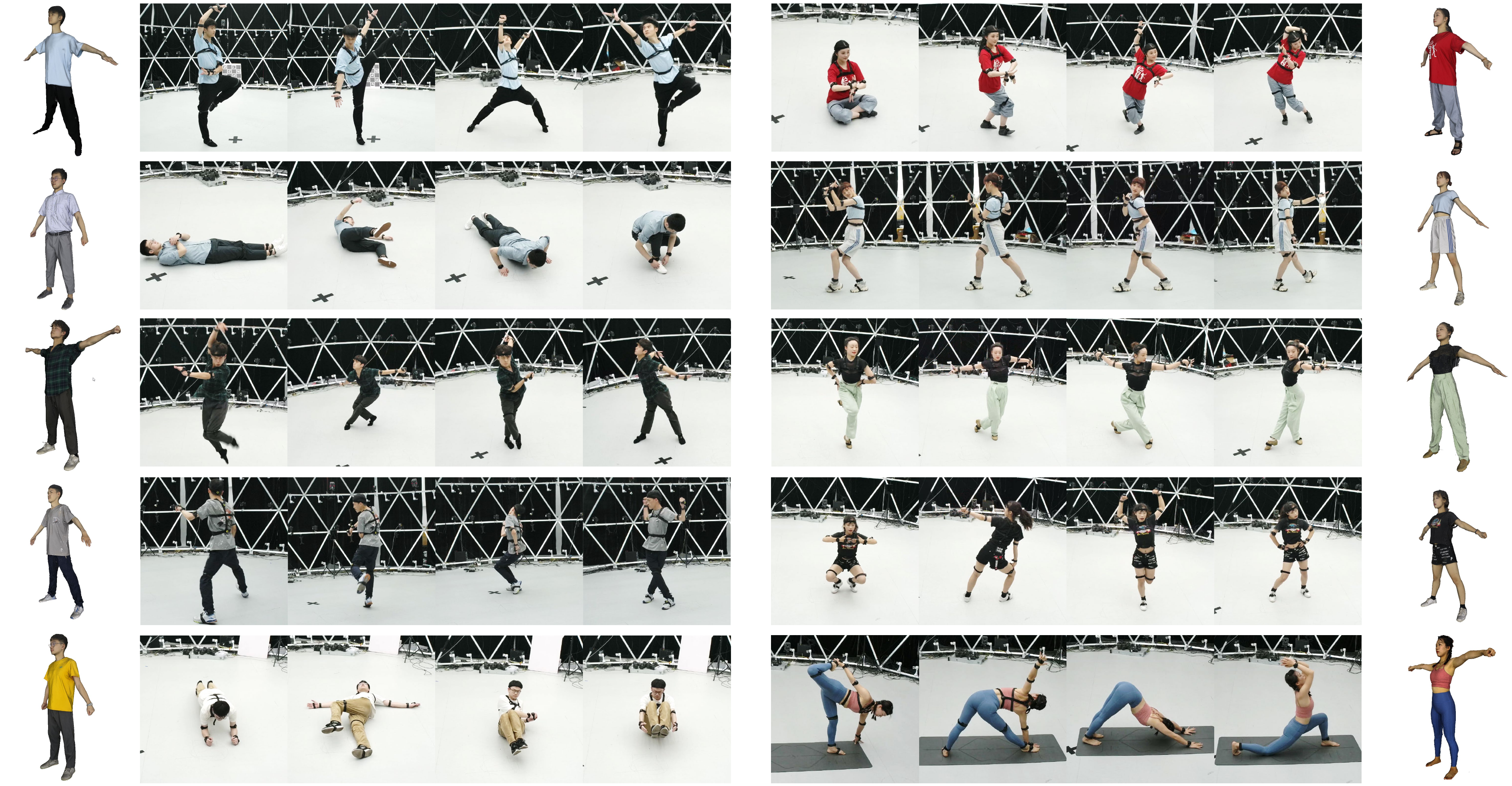}
	\caption{HCM dataset contains a wide variety of challenging motions along with abundant images, inertial measurements, and pre-scanned templates.}
	\label{fig:dataset}
    \vspace{3mm}
\end{figure*}

\begin{table*}[th]
	\begin{center}
		\centering
		\resizebox{1\textwidth}{!}{
    		\begin{tabular}{llcccccccccc}
    			\hline
    			Modality& \multicolumn{1}{l}{Dataset} & Daily Clothing  & Actor Scans & Music  & Views        & IMUs       & Minutes      & Poses & Images  & Inertia   \\
    			\hline
    			\multirow{5}{*}{Single}&  AIST++~\cite{li2021learn} & \checkmark & - & \checkmark & 9         & -      & 239       & 860K   & 7.7M & - \\
    			& Human3.6M~\cite{ionescu2013human3} &  \checkmark &  \checkmark & - & 4             & -    & 300      & 900K    & 3.6M & - \\
    			&  MPI-INF-3DHP~\cite{mehta2017monocular} &  \checkmark & -  & -  & 14             & -    & 26   & 93K    & 1.3M & - \\
    			&  LSP~\cite{johnson2010clustered} &  \checkmark & - & -  & 1                & - & -  & 12K   & 0.012M  & - \\
    			&  DIP-IMU~\cite{huang2018DIP} &  - & -  & -   & -      & 17        & 90     & 324K  & -  & 5.5M  \\
    			\hline
    			
    			\multirow{3}{*}{Hybrid}&  TotalCapture~\cite{TotalCapture}    &  - & - & -  &      8   & 13   & 49       & 176K    & 1.4M    & 2.3M    \\
    			&  3DPW~\cite{vonMarcard2018} &  \checkmark & \checkmark & - &   1          &  $6 \sim17 $      & 11      & 37K  & 0.037M  & 0.33M \\
    			\rowcolor{black!7}  &  HCM Dataset (ours) &  \checkmark & \checkmark   & \checkmark &  \textbf{14}          & \textbf{17}      & 176       & 633K & \textbf{8.9M}  & \textbf{10.8M}   \\
    			\hline
    		\end{tabular}
        }
		\caption{HCM dataset is a new largest multi-modal human motion dataset recorded from the most camera views and IMUs, providing pre-scanned templates of daily clothing actors, paired music, and the most images and inertial measurements.
		}
		\label{tab:dataset}
		\vspace{2mm}
	\end{center}
\end{table*}

\subsection{Sensor Calibration}
Before capturing, we first calibrate the IMUs and camera by solving rotation transformations $\mathbf{R}_{I2C}$ and $\mathbf{R}_{S2B,n}$, where $\mathbf{R}_{I2C}$ is the transformation between inertial frame $\mathcal{F}_I$ and camera frame $\mathcal{F}_C$, and $\mathbf{R}_{S2B,n}$ is the transformation from the $n$-th IMU sensor $\mathcal{F}_{\textbf{s}_n}$ to $\mathcal{F}_{\textbf{b}_n}$ of its corresponding bone $\textbf{b}_n$. 
With only one camera, it is difficult to calibrate by continuous per-frame optimization~\cite{Zheng2018HybridFusion} on a complicated motion sequence. 
Instead, we adopt a two-frame calibration strategy, one frame for A-pose and the other one for T-pose.

Denote orientation observations of IMUs and bones (obtained from pre-defined calibration pose) as $\mathbf{\widetilde{R}}_n^{(A)}$, $\mathbf{\widetilde{R}}_n^{(T)}$ and $\mathbf{\hat{R}}_{\textbf{b}_n}^{(A)}$, $\mathbf{\hat{R}}_{\textbf{b}_n}^{(T)}$ respectively. Thus both observation pairs of each IMU sensor can be unified by $\mathbf{R}_{I2C}$ and $\mathbf{R}_{S2B,n}$ under $\mathcal{F}_C$. Therefore, $\mathbf{R}_{I2C}$ and $\mathbf{R}_{S2B,n}$ can be determined by solving following systems:

\begin{align}
    \mathbf{R}_{I2C}\mathbf{\widetilde{R}}_n^{(A)} = \mathbf{\hat{R}}_{\textbf{b}_n}^{(A)}\mathbf{R}_{S2B,n}\\
    \mathbf{R}_{I2C}\mathbf{\widetilde{R}}_n^{(T)} = \mathbf{\hat{R}}_{\textbf{b}_n}^{(T)}\mathbf{R}_{S2B,n}
\end{align}

\begin{figure*}[t]
	\centering
	\includegraphics[width=1\linewidth]{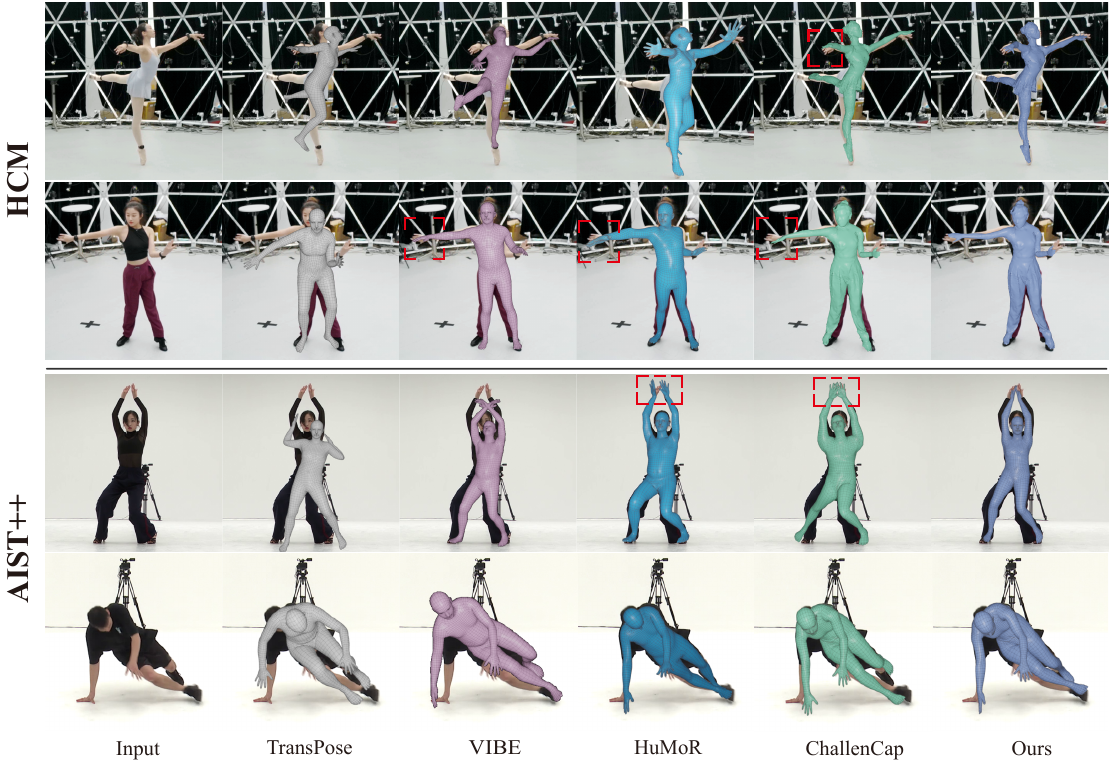}
	\caption{More comparisons that show the superior of HybridCap on the performance of both overlay and limb orientation.}
	\label{fig:comp}
	\vspace{-2mm}
\end{figure*}

\vspace{-2mm}
\subsection{Acceleration Synthesis}
For AIST++ \cite{li2021learn, aist-dance-db} dataset whose inertia is unavailable, we synthesize the needed IMU observations as input using SMPL \cite{SMPL2015} mesh, which is recovered from pseudo ground-truth SMPL parameters. The formulation is as following. 
\begin{equation}
    \mathbf{A}_n^{(t)} = (\mathbf{P}_n^{(t+1)} - 2\mathbf{P}_n^{(t)} + \mathbf{P}_n^{(t-1)}) / st^2
    \label{eq:acceleration}
\end{equation}
where $\mathbf{P}_n$ denotes the simulated IMU positions corresponding to some predefined vertices of SMPL mesh.
And we use smooth factor of $n=4$ as same as \cite{TransPose2021}, since it produces accelerations most close to real sensor measurements. 

\vspace{-2mm}
\subsection{Training Details}

We train our hybrid motion inference module with Adam optimizer \cite{kingma2014adam} using multi-phase strategy. In the first phase, we train limb tracker using $\mathcal{L}_{limb}$ for 20 epochs, and then we add $\mathcal{L}_{body}$ to train body tracker for 20 epochs. In the second phase, we pretrain the hybrid IK solver and the root tracker for 10 epochs using $\mathcal{L}_{prior}$ and $\mathcal{L}_{trans}$. In the last phase, we add full $\mathcal{L}_{IK}$ to train the whole network for 100 epochs. 
In our experiments, $\mathcal{L}_{limb}$, $\mathcal{L}_{body}$ and $\mathcal{L}_{IK}$ are weighted equally. In $\mathcal{L}_{IK}$ we set $\lambda_{\mathrm{2D}}=1$, $\lambda_{\mathrm{acc}}=10$ and $\lambda_{\mathrm{ori}}=30$, while $\lambda_{\mathrm{prior}}=0.01$ and $\lambda_{\mathrm{trans}}=0.01$.
During training, two Nvidia RTX3090 GPUs are utilized. We set the dropout ratio as 0.5 for all the LSTM layers. The batch size is 16 and the sequence length is 360, while the learning rate is $1\times10^{-4}$, the decay rate is 0.1 (final 50 epochs).

\vspace{-2mm}

\begin{figure*}[t]
	\centering
	\includegraphics[width=1\linewidth]{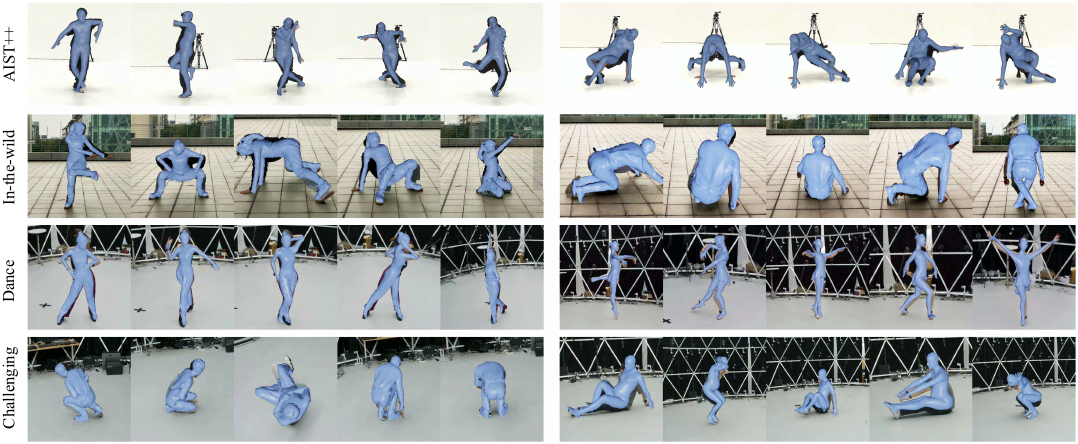}
	\caption{More qualitative results that show the robustness of HybridCap capturing the challenging motions.}
	\label{fig:gallery}
\end{figure*}

\subsection{HCM Dataset}

Since accurate 3D pose ground truth annotations are highly expensive and tedious, as an alternative, weak supervision from a large amount of relatively affordable sensors in our training mechanism is critical for performance improvement as pointed out in \cite{DeepCap_CVPR2020}. 
There are existing datasets for 3D pose estimation and human motion capture, as Tab.~\ref{tab:dataset} summarizes.
However, they do not satisfy the requirements about sufficient camera views, actor-mounted IMUs, challenging motions and person-specific template that are critical for our multi-modal supervision and evaluation.
Thus, to thoroughly evaluate our method, we build and propose a new dataset Hybrid Challenging Motions (HCM). 

\myparagraph{Data Collection.}
We capture the motion data using a hybrid capture studio in an 8m-diameter multi-camera dome, where synchronized 14 RGB cameras~\cite{Z-CAM} and 17 actor-mounted IMUs~\cite{NOITOM} are used. 
The capture studio is shown as Fig.~\ref{fig:wear}. The studio covers a wide range of viewpoints with 6 cameras mounted at chest height with a roughly $15^{\circ}$ elevation variation similar to \cite{mehta2017monocular}, and another 8 higher and angled down $45^{\circ}$ cameras with a roughly $5^{\circ}$ elevation variation.
Besides, the positions of actor-mounted IMUs cover most human rigid body parts, which in order are: head, left shoulder, right shoulder, left upper arm, right upper arm, left lower arm, right lower arm, left hand, right hand, back, hip, left upper leg, right upper leg, left lower leg, right lower leg, left foot and right foot. 
This dense multi-modal capture setting provides maximized viewpoint variation and inertial observations of moving rigid body parts, which is necessary for our dense weak supervision and thorough evaluation.

\myparagraph{3D Reconstruction.}
Before recording, We scan the actor with a 3D body scanner to generate the textured template mesh of the actor. Then, we rig it automatically by fitting the Skinned Multi-Person Linear Model (SMPL)~\cite{SMPL2015} to the template mesh and transferring the SMPL skinning weights to the scanned mesh.
Note that we follow \cite{EventCap_CVPR2020, challencap} to transfer the skinning weights to maintain compatibility with SMPL's skeleton. It would not bias the quantitative results since we use SMPL's joints as 3D keypoints for regression or reprojection. 
Then we calibrate the cameras using Zhang's method~\cite{zhang2000flexible} and the IMUs using pre-definition and optimization similar to \cite{HybridFusion}.
After recording, the 3D joint locations are triangulated from the multi-view human pose 2D keypoints extracted by OpenPose~\cite{OpenPose}. 
During the triangulation phase, we apply temporal smoothness and bone length constraints to improve the quality of the reconstructed 3D joint locations.
Finally, we fit SMPL skeleton to the 3D joint locations and IMU measurements to obtain the ``pseudo ground-truth'' reference SMPL pose parameters along with translation and refined 3D joint locations.

\myparagraph{Dataset Description.}
HCM is a large-scale 3D human motion dataset that contains 16 subjects and over 100 one-minute sequences with a wide range of challenging motions. 
It has the following extra data for each frame:
\begin{itemize} 
\setlength\itemsep{0em}
    \item 3D pre-scanned template of actor;

	\item Images from 14 corresponding views of camera;

	\item Raw measurements from 17 actor-mounted IMUs;

	\item 14 views of camera intrinsic and extrinsic parameters;
	
	\item 17 IMUs calibration parameters;
	
	\item 25 OpenPose-format human joint locations in both 2D and 3D;
	
	\item 24 SMPL pose parameters along with translation and 10 neutral SMPL shape parameters;
	
	\item Paired music if it is a dance motion sequence.

\end{itemize}

To our knowledge, HCM is the largest multi-modal 3D human motion dataset containing a wide variety of dance and rare challenging motions. 
Tab.~\ref{tab:dataset} provides the statistics about the human motion datasets widely used in the community. 
Our HCM dataset recorded from both the most camera views and IMUs provides 8.9M images and 10.8M inertial measurements. 
And it contains the pre-scanned templates of the actors, which provides the accurate anthropometry and enable future research direction to capture not only rigid body motion but also non-rigid surface deformation.
Note that the pre-scanned template is not a prerequisite for our approach, and instead actor-specific key bone lengths are what we need. One can use image-based estimation algorithms to obtain a template SMPL mesh with fixed bone lengths, as shown in the results on AIST++ and 3DPW. Thus, scanning is optional and we adopt it only for the visualization compared to the template-based method~\cite{challencap}.
Besides, we do not use the TotalCapture~\cite{TotalCapture} dataset because of its unusual appearance of the actors in the professional mocap suits, which bring domain gap against daily clothing in terms of 2D features.

HCM contains 12 dance genres: folk dance (Han folk, Dai folk), ballet, Latin (Rumba, Cha-cha, Samba, Tango, Jive), Waltz, Jazz, Hip-hop, and cheerleading. And each dance sequence is paired with synchronized music, making it useful for other research directions such as dance generation.
It further contains a large amount of rare challenging motions, such as rolling, tumbling, fitness, boxing, yoga and forth on. 
As shown in Fig.~\ref{fig:dataset}, a gallery sampled from HCM dataset is provided.

\vspace{-1pt}
Besides, for qualitative evaluation, we recorded several indoor and in-the-wild one-minute challenging motion sequences using a single camera and 4 IMUs mounted on lower arms and lower legs. 
More comparisons and qualitative results are shown in Fig.~\ref{fig:comp} and Fig.~\ref{fig:gallery} respectively.

\end{document}